\documentclass[mlabstract,twocolumn]{jmlr}

\usepackage{longtable}
\usepackage{booktabs}
\usepackage[load-configurations=version-1]{siunitx} %

\theorembodyfont{\upshape}
\theoremheaderfont{\scshape}
\theorempostheader{:}
\theoremsep{\newline}

\jmlrvolume{}
\firstpageno{1}
\editors{}
\jmlryear{2022}
\jmlrworkshop{Machine Learning for Health (ML4H) 2022}
\title[Quantifying Token-level Sensitivity for Readmission Prediction]{Language Model Classifier Aligns Better with Physician Word Sensitivity than XGBoost on Readmission Prediction}

\author{
\Name{Ming Cao\nametag{\thanks{These authors contributed equally}\thanks{NYU Center for Data Science}\thanks{NYU Langone Health}}}\Email{mc7787@nyu.edu}
\AND 
\Name{Grace Yang\nametag{\footnotemark[1]\footnotemark[2]\footnotemark[3]}} \Email{gy654@nyu.edu} 
\AND
\Name{Lavender Y. Jiang\nametag{\footnotemark[2]\footnotemark[3]}}\Email{lyj2002@nyu.edu}
\AND
\Name{Xujin C. Liu\nametag{\footnotemark[3]\thanks{NYU Tandon School of Engineering}}} \Email{chris.liu@nyu.edu}
\AND
\Name{Alexander T. M. Cheung\nametag{\footnotemark[3]\thanks{NYU Grossman School of Medicine}}}\Email{alexander.cheung@nyulangone.org}
\AND
\Name{Hannah Weiss\nametag{\footnotemark[3]}} \Email{hannah.weiss@nyulangone.org}
\AND
\Name{David Kurland\nametag{\footnotemark[3]}} \Email{david.kurland@nyulangone.org}
\AND
\Name{Kyunghyun Cho\nametag{\footnotemark[2]\thanks{Prescient Design}}} \Email{kc119@nyu.edu}
\AND
\Name{Eric K. Oermann\nametag{\footnotemark[3]\footnotemark[5]\footnotemark[2]}}\Email{eric.oermann@nyulangone.org}
}

\begin{document}

\maketitle

\begin{abstract}
Traditional evaluation metrics for classification in natural language processing such as accuracy and area under the curve fail to differentiate between models with different predictive behaviors despite their similar performance metrics. We introduce sensitivity score, a metric that scrutinizes models' behaviors at the vocabulary level to provide insights into disparities in their decision-making logic. We assess the sensitivity score on a set of representative words in the test set using two classifiers trained for hospital readmission classification with similar performance statistics. Our experiments compare the decision-making logic of clinicians and classifiers based on rank correlations of sensitivity scores. The results indicate that the language model's sensitivity score aligns better with the professionals than the xgboost classifier on tf-idf embeddings, which suggests that xgboost uses some spurious features. Overall, this metric offers a novel perspective on assessing models' robustness by quantifying their discrepancy with professional opinions. Our code is available on GitHub.\footnote{\url{https://github.com/nyuolab/Model_Sensitivity}}\end{abstract}
\begin{keywords}
hospital readmission prediction, sensitivity analysis, model interpretability, MIMIC-III
\end{keywords}

\section{Introduction}
\label{sec:intro}
Predicting 30-day all-cause hospital readmission is a classical problem in medical informatics as 30-day readmissions are associated with longer hospital stays, higher mortality rates, and significant operating expenses. Several natural language processing (NLP) tasks have used models based on BioClinicalBERT, bag-of-words, and BI-LSTM; however, they all achieve similar test performances despite difference in modelling techniques \citep{jamei_predicting_2017,van_walraven_lace_nodate, xiao_readmission_2018, alsentzer_publicly_2019,huang_clinicalbert_2020}. It is challenging to select the best model for deployment in such cases with similar performance statistics, particularly in the case of complex machine-learning models whose decision-making logic may differ substantially from professionals. This lack of interpretability and insight into model predictions is a substantial barrier to medical deployment for the fear of potentially causing harm or incurring additional costs \citep{jackson_agile_2019, xiao_opportunities_2018}.

To keep the model accountable, we propose a general-purpose evaluation framework to quantify the classifier's sensitivity to individual words based upon text perturbations. Text perturbations have been used for model bias detection \citep{prabhakaran_perturbation_2019}, but our goal is to recast it for model sensitivity quantification as a function of model vocabulary. Similar attempts to evaluate models have been proposed, but with their respective limitations. We discuss the existing methods and the motivation to create our metric in Appendix \ref{sec:formalization}. For a particular model, we compute the sensitivity score for each target word as the $L_1$ distance between the classifier outputs before and after perturbations of the word in the test corpora. We test the language model classifier versus an extreme gradient boosted tree (xgboost) classifier \citep{chen_xgboost_2016} with tf-idf (Term Frequency-Inverse Document Frequency) embeddings as features \citep{sparck_jones_statistical_1972}.

The sensitivity score quantifies the significance of tokens as perceived by the model, which enables us to inspect the classifier's decision-making logic in the context of professional opinions. Ideally, if the classifier is sensitive to clinically significant words deemed by professionals, and indifferent to words considered trivial, then the classifier conforms to professional opinions and is less likely to have learned spurious correlations. For example, a reliable readmission classifier could be sensitive to words like ``dementia" and indifferent to words like ``parent". Sanity checks like this example enables us to rule out bias and lend credence to the model. 

The major contributions of this abstract is: we offer a novel perspective of evaluating the model's accountability. Models could make correct predictions simply by exploiting spurious patterns from the data. Such models are susceptible to distribution shifts, and therefore less reliable in a changing clinical environment once deployed. Traditional metrics typically do not examine which features models use to make predictions, but rather focus on evaluating performance statistics. Thus, it is necessary to zoom into the model's behavior at the vocabulary level. A divergence between clinicians and the model in word significance rankings could indicate liability to distribution shifts or potential insights overlooked by humans.

\section{Methods}
\label{sec:method}


In our study, we test the sensitivity of two text-based readmission predictors: a finetuned BioClinicalBERT and an xgboost on tf-idf embeddings. These models' overall performances are evaluated by their AUROC (The area under the receiver operating characteristic) and AUPRC (The area under the precision recall curve) scores on the test dataset. Their accountabilities are evaluated by the correlation of their sensitivity rankings on a set of 49 hand-selected words with those of physicians.



\subsection{Dataset}
Both of the models are trained on a readmission dataset derived from MIMIC-III, a database of electronic health records from ICU patients at the Beth Israel Hospital \citep{johnson_mimic-iii_2016}. Each note  has  a binary label for readmission. A positive label suggests readmission within 30 days following the patient's discharge. The labeled dataset has 6\% positive labels, with 52,725 examples and a 70\% train, 15\% validation and 15\% test split. For more details see Appendix \ref{appen:data}.

For assessing model robustness and generalizability, a second readmission dataset is obtained from the NYU Langone Health System as part of an IRB approved study of inpatient readmissions. The NYU readmissions dataset consists of 45,120 examples from all clinical departments with 10.67\% of positive labels. 

\subsection{Readmission Classifier}

\textbf{Language-model based classifier}: BioClinicalBERT is a transformer encoder model pretrained on PubMed abstracts and MIMIC-III. It uses the original vocabulary of BERT and a Wordpiece tokenizer \citep{song_fast_2021}.

We finetune the pretrained model on our readmission dataset for 10 epochs and searched for a learning rate that gives the lowest validation loss. We use weighted cross entropy due to label imbalance. See appendix \ref{appen: imbalance} for more details.




For inference, we use a threshold of 0.35 to convert the redicted probabilities to binary labels (label 1 is assigned if the model’s predicted probability is above 0.35 and vice versa) such that we reach 70\% recall on the validation set.



\textbf{Baseline Model (tf-idf+xgb):} 
As a baseline model, we build a xgboost classifier using tf-idf embeddings as features. We select xgboost to compare the language model based classifier with a traditional machine learning classifier. See Appendix \ref{appen: B.2} for more details.

Tf-idf has a less informative embedding, but is faster to train. The tf-idf based xgboost model does not use self-attention and positional embedding to incorporate semantic context and word orders into the representation. This makes its embeddings simply reflect how important each word is to a note compared to the entire corpus. On the other hand, tf-idf has a linear computational complexity for training with respect to the input sequence length, whereas attention-based model such as BioClinicalBERT has quadratic complexity.

\subsection{Metric: Token Sensitivity Score}

We present the token \textbf{sensitivity score}, a metric to gauge the difference between classifier outputs before and after text perturbations. We say a classifier is sensitive to a token if the probability output changes notably after we perturb that token.  

In this work, we explore swapping words as the perturbation. In our experiment, we used 3 types of perturbations: the uniform perturbations, the 1-gram perturbations, and the context perturbations. The uniform perturbation replaces a token of interest with another token uniformly sampled from the vocabulary of BioClinicalBert. The 1-gram perturbation replaces a token of interest with one of the five most frequent tokens. The context perturbation replaces a token of interest with one of the five most likely words according to BioClinicalBert's masked language modeling.

To illustrate the intuition of sensitivity score, consider determining whether or not a patient needs to be hospitalized. If ``the patient has stroke", then we think it's likely that the patient is hospitalized. After context perturbation, if the text is swapped to ``the patient has flu", then the patient is probably fine. Since the swap changes our opinion significantly, we say ``stroke" is a sensitive word for determining hospitalization. Now we perturb the statement by swapping ``has" with ``got". In this case, the semantic meaning does not change much, and we say ``has" is a relatively nonsensitive word for determining hospitalization.

In order to compare the model's behaviors with human doctors, we further propose the token \textbf{sensitivity score rank}. For each token, we compute the token's rank in terms of sensitivity scores among all the target words. We can then use the correlation between a model's rank and human's rank to assess how much a model's decisions align with human doctors' beliefs. For more details on the mathematical formulation, see the remaining subsections.

\subsubsection{Notations}
To formalize the token sensitivity score, we need to first introduce some notations. A classifier $f$ is a function that takes in a note $x$ comprising of a set of tokens sampled from a vocabulary and outputs a probability $p$. For example, a constant classifier that always  predict 30-day readmission using medical discharge summaries is $f_{\text{const}}(x)=1$.  

A note $x$ is a sequence of $n$ tokens/words $(w_1,w_2,\dots,w_n)$. For example, a note could be (``his", ``mom", ``visited").

We can perturb a note with a \textbf{perturbation function} $g$, which is parameterized by a token of interest $u$ and a perturbation filter $h$. The perturbation function $g$ replaces the first occurrence of the token of interest $u$ with a perturbed token given by the filter $h$. We only replace the first occurrence (denoted as one-swap) to reduce the impact of token frequency, which positively correlates with changes in predicted probabilities as shown in Appendix \ref{appendix:B.1}. 

For example, if we are interested in how sensitive a classifier $f$ is to seeing ``dad" rather than ``mom", then our word of interest is $u=\text{``mom"}$, our perturbation filter is $h(\text{``mom"})=\text{``dad"},$ and a perturbation example is:
\begin{align*}
g_{u,h}(&\text{(``his",``mom",``visited")})\\
=&\text{(``his",``dad",``visited")}.
\end{align*}
More generally,
\begin{align*}
    g_{u,h}(x)&=g((w_1,\dots,w_n))\\
    &=(w_1,\dots,w_{k-1},h(w_k),w_{k+1},\dots,w_n),
\end{align*}

where $k$ is the first location where the word of interest $u$ appears. Note that for a fixed perturbation function $g_{u,h}$, the input  $x$ must contain $u$.  Otherwise, the perturbation function does not change the note.

\subsubsection{Quantifying Sensitivity w.r.t. Perturbation Function}

With a perturbation function $g_{u,h}$, we can quantify the sensitivity as the difference in predicted readmission probabilities before and after perturbing the token of interest $u$ with the filter $h$, as measured by $L_1$ distance:
$$d_{f}(g_{u,h})(x)=|f(x)-f(g_{u,h}(x))|$$
For example, given note $x=$(``his", ``mom", ``visited"), the example classifier $f_{\text{const}}$, and the perturbation function $g_{u,h}$ defined in section \ref{sec:formalization}, we have 
$$d_{f_{\text{const}}}(g_{u,h})(x)=|1-1|=0,$$
since the constant classifier always predicts positive labels regardless of the input text.

\subsubsection{Averaging across Different Perturbations and Notes}

We are interested in a general-case estimate of the sensitivity of a classifier with respect to perturbing a token. This motivates us to look at the difference in predicted probability across different perturbations and in different notes. For example, the change (``his mom visited"$\to$``his dad visited") might lead to a smaller difference than (``his mom is pregnant"$\to$``his dad is pregnant") because the semantic change is less substantial.

To approximate the sensitivity of classifier $f$ with respect to various perturbations, we consider a set of perturbation filters $H=\{h_1,\dots,h_m\}$. The choice of $H$ is up to the users. We use uniform filters to introduce randomness in perturbations. To limit the degree of perturbations out of the training distribution, we add filters based on distributions induced from the training set, i.e., the 1-gram distribution of a well-trained masked language model. 

We define the \textbf{note-level sensitivity score} with respect to a token $u$ in a note $x$ as the average difference in predicted probabilities after applying a filter in $H$.
\begin{equation*}
    \overline{{d_{f}}}(g_{u,H})(x) = \frac{1}{m}\sum_{i=1}^m d_{f}(g_{u,h_i})(x)
\end{equation*}

To approximate the sensitivity across various notes, we consider a set of notes $X=\{x_1,\dots,x_{l}\}$ and define the \textbf{overall sensitivity score} with respect to a token $u$ as the average of note-level sensitivity score:
\begin{equation*}
    \overline{{d_{f}}}(g_{u,H})(X) = \frac{1}{l}\sum_{j=1}^l\overline{{d_{f}}}(g_{u,H})(x_j)
\end{equation*}

\subsubsection{Comparing Sensitivity of Different Models}
Different models $f_1, f_2$ have respective ranges of output probabilities, making it unfair to directly compare the changes in predicted probabilities. For example, $f_1$ could mostly predict $p\in[0.1,0.3]$, whereas $f_2$ predict $p\in[0.4,1]$. In this case, comparing the overall sensitivity scores of $f_1$ and $f_2$ is biased towards the conclusion that $f_2$ is more sensitive, because its range of outputs is higher. 


To address this issue, we compare relative sensitivity as opposed to the absolute sensitivity. That is, we want to say whether $f_1$ is \textit{more} sensitive to $u$ compared to $f_2$.

We measure relative sensitivity with the sensitivity rank. Given a set of token of interest $U$, we calculate $f$'s sensitivity scores of each token within it. A token $u$ has a sensitivity rank of $r_{u,f}$ if it has the $r_{u,f}$-th highest sensitivity score among $U$. 

Now we can compare relative sensitivity of different classifiers using the sensitivity ranks. For example, we say $f_1$ is more sensitive to $u$ than $f_2$ if $r_{u,f_1}<r_{u,f_2}$.

\subsubsection{example with mimic-iii}
\label{sec: example with mimic-iii}
We choose 49 tokens of interest to evaluate our two readmission classifiers, $f_{\text{BioClinicalBERT}}$ and $f_{\text{tfidf}+\text{xgb}}$.  For each token $u$ of interest, we calculate the overall sensitivity score $\overline{{d_{f}}}(g_{u,H})(X)$, with $H$ as one of the fifteen perturbation filters, and $X$ as the subset of our dataset that contains the token $u$. Our perturbation filters $H$ is partitioned into 3 sets of 5 replacement filters: the uniform perturbations, the 1-gram perturbations, and the context perturbations. For more details check Appendix \ref{appen:C2}.

\vspace{-2mm}
\section{Experiment Results}

\textbf{Tf-idf+xgb has slightly better AUC than BioClinicalBERT.}  As shown in \autoref{tab:language_model}, The standard deviation of BioClinicalBERT is 2.77 times that of tf-idf+xgb in AUROC, and 3.25 times that of tf-idf+xgb in AUPRC. While the marginal advantage of tf-idf+xgb suggests that word order and semantic context are not crucial, our next result shows that tf-idf+xgb's predictive behaviors align worse with the physicians. 

\begin{table}[h]
\small
\centering
  {\begin{tabular}{lll}
  \toprule
  \bfseries Model & \bfseries AUROC & \bfseries AUPRC\\
  \midrule
bioclinical & 0.6995$\pm$0.0036 & 0.1224$\pm$0.0065 \\
tf-idf+xgb & \textbf{0.7150}$\pm$0.0013 & \textbf{0.1340}$\pm$0.0020 \\
  \bottomrule
  \end{tabular}}
  \caption{Comparison of readmission AUC between BioClinicalBERT and tfidf+xgb. Test statistics are from 5 trials with distinct random seed. }\label{tab:language_model}
  \vspace{-6mm}
\end{table}

  


\textbf{Language-model based readmission classifier correlates better with clinicians' sensitivity.} To investigate which model is more reliable, we select 49 target words based on professional inputs and collect the sensitivity rankings of 3 readmission predictor: the finetuned BioClinicalBERT, tfidf+xgb, and 3 human clinicians. We invite three clinicians to rank 49 target words with a score from 1 to 5 to reflect the significance of each word in readmission prediction. A higher rank (smaller number) indicates a more important word for decision making. We obtain the overall clinician rankings by averaging the ranks across all three clinicians. We then assess the model's similarity to professional judgements through the Spearman rank correlation \citep{spearman_proof_1904} between models' rankings and the overall clinician ranking.  \autoref{tab:corr} shows that BioClinicalBERT has a higher rank correlation, despite a slightly lower AUC in \autoref{tab:language_model}. Check Appendix \ref{appen:C3} and Appendix \ref{apd:second} for more details.

\begin{table}[hbtp]
\centering
  {\begin{tabular}{ll}
  \toprule
  \bfseries classifier & \bfseries rank\_correlation \\
  \midrule
BioClinicalBERT & \bf{0.5754}\\
tf-idf+xgb & 0.1259 \\
  \bottomrule
  \label{table:rank_correlation}
  \end{tabular}}
  \vspace{-6mm}
  \caption{Spearman rank correlation between the two classifiers' rankings and the physicians' ranking. }\label{tab:corr}
  \vspace{-6mm}
\end{table}

In addition to the quantitative difference in Spearman rank correlations, to understand the difference between BioClinicalBERT and tfidf+xgb, we perform a qualitative analysis on where these models disagree based on \autoref{tab:49_words_1} and \autoref{tab:49_words_2}. Words like ``tumor", ``pancreatic", and ``dementia" are considered important by clinicians and BioClinicalBERT but neglected by the tfidf+xgb model. Meanwhile, words like ``increase", ``prescribed", and ``blood" are considered important by the tfidf model but trivial by the other two. The inconsistency between the rankings of clinicians and the tfidf+xgb model shows its potential reliance on spurious features, rendering it less robust.

To empirically verify that the tfidf+xgboost model is less robust, we make zero-shot inferences on a held-out readmission dataset from NYU Langone Health. The result in \autoref{tab:nyudata} partially verifies the legitimacy of our sensitivity metric's ability in evaluating robustness.

\begin{table}[hbtp]
\small
\centering
  {\begin{tabular}{lll}
  \toprule
  \bfseries Model & \bfseries AUROC   & \bfseries AUROC\_drop \\
  \midrule
    bioclinical & 0.633 & \textbf{0.0665} \\
    tf-idf+xgb & 0.605 &  \textbf{0.11}  \\
  \bottomrule
  \end{tabular}}
  \caption{tfidf+xgboost has a greater drop in AUROC than BioClinicalBERT when inferencing on new data }\label{tab:nyudata}
  \vspace{-10mm}
\end{table}

\section{Discussions}
\textbf{Limitation:} Our metric has a large spatial and computational complexity.  Generally, the complexity is worse than $NM$, where $N$ is the dataset size and $M$ is the number of perturbations. We can reduce complexity with the Monte Carlo method: subsampling the dataset and the perturbation filters. Methods of improving computational inefficiency is a direction of future research.

\textbf{Implications:} BioClinicalBERT might use the semantic context to extract more holistic information from a patient, as opposed to relying on statistical correlations that do not apply to specific subgroups. For example, an ``increase" of blood pressure might be dangerous for patients with heart problems, but is a sign of recovery for patients with low blood pressure.



\newpage

\acks{We would like to acknowledge Michael Costantino, Ph.D. and Kevin Yie, M.S., from the NYU Langone High Performance Computing (HPC) team. Without their tireless assistance in building and maintaining our GPU cluster, none of this research would have been possible. We would also like to thank Ben Guzman from the NYU Langone Predictive Analytics Unit and Vincent J. Major from NYU Grossman School of Medicine for their help with learning the SQL data structures used as part of this work. This work is supported in part by NSF under grants 1922658. Additional funding comes from NYU Grossman School of Medicine.}

\bibliography{pmlr-sample}

\appendix
\section{Data}
\label{appen:data}

\subsection{Data Source}
We use three tables from the MIMIC-III in our study: patients, admissions, and noteevents. The three tables have ``subject\_id", ``hadm\_id", and ``row\_id" as their respective primary keys. We use the discharge note as our input since it contains the most information as a summary of the entire visit.


\subsection{Label Generation}

For each distinct discharge note associated with an encounter, we generate a binary readmission label based on the patient's medical record. A positive label suggests readmission within 30 days following the patient's discharge. 

We generate the readmission labels as follows: for each patient, we order their encounters by admission time. For each of their encounters, we calculate the readmission interval between the discharge time for the current visit and the subsequent admission time. If an encounter does not have subsequent admissions or the readmission interval is longer than 30 days, we assign a negative label. Otherwise, we assign a positive label.

To properly handle boundary cases, we only consider encounters that are discharged at least a month before the latest admission time. This prevents false negative labels with unobserved readmissions outside the dataset.

\section{Training Details}
\subsection{Finetuning BioClinicalBERT}
\label{appen: imbalance}
In finetuning, we search learning rate from \{2e-5, 2e-6, 2e-7, 2e-8, 2e-9\} using ray-tune \citep{liaw2018tune}. For each learning rate, we finetune the model for 10 epochs using a Nvidia-3090 GPU for around 140 minutes. We select the model with best validation loss, which uses a learning rate of 2e-5 and 5 epochs. 

To address the label imbalance issue, we use a weighted cross entropy loss function to increase the penalty for misclassifying the minority class. Specifically, we weigh the positive example with the ratio of negative examples in the entire dataset; Similarly, we weigh the negative example with the ratio of positive examples.

\subsection{Training tf-idf+xgb}
\label{appen: B.2}

The model first convert the input texts to a word-count vector, then calculate the tf-idf embedding. Next, xgboost uses this embedding matrix for binary classification. We repeat the training five times with distinct random seeds (24, 42, 61, 67, 70) for a total of three minutes.

\section{Figures}\label{apd:first}

\begin{figure}[ht]
    \centering
    \includegraphics[width=\linewidth]{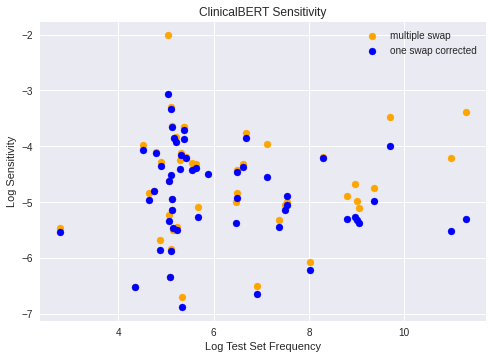}
    \caption{Comparison of BioClinicalBERT word sensitivities using the one-swap and multiple-swap scheme on readmission prediction. The one-swap scheme counteracts the effect of frequency on sensitivity score.}
    \label{fig:bs}
\end{figure}

\begin{figure}[ht]
    \centering
    \includegraphics[width=\linewidth]{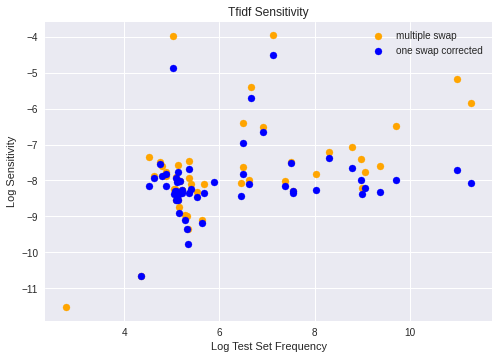}
    \caption{Comparison of tf-idf word sensitivities using the one-swap and multiple-swap scheme on readmission prediction. The one-swap scheme counteracts the effect of frequency on sensitivity score. }
    \label{fig:ts}
\end{figure}

\section{Metric Formalization}\label{sec:formalization}
\label{metric_formalization}

\subsection{Previous Works}
\label{sec:existing-metrics}Existing metrics for model interpretation mainly focus on unraveling individual predictions or mimicking local behaviors of the model. However, these methods have two limitations: first, they focus on attributing decision making to local features in individual examples; second, they cannot quantify explainability as measured by alignment with some ground-truth references. For example, SHAP (Shapley Additive exPlanations) \citep{lundberg_unified_nodate} may be able to tell us that “alcohol” is important for predicting that an individual depressed patient would be readmitted. But SHAP cannot tell us whether “alcohol” is a key feature for predicting readmission over an entire readmission dataset (not necessarily the case, because the use of “rubbing alcohol” indicates good hygiene). Further, if SHAP tells us that “alcohol” is important for an individual prediction, we cannot interpret this finding without checking its alignment with some experts. One way is to consult some physicians: “Is it good that my model thinks alcohol is an important feature?” Our proposed sensitivity score fills this gap by: first, quantifying the sensitivity of each token over an entire dataset (e.g., overall, “alcohol” is not a sensitive feature); second, quantifying the alignment of model’s sensitivity with a reference (e.g., overall, language model’s sensitivity to words are more similar to trustworthy physicians than tf-idf models).

\section{Derivation details}

\label{appendix:B}
\subsection{Reasons for Substituting the First Occurrence}
\label{appendix:B.1}

This section explains the rationale behind word substitution only once regardless their total occurrences. 

Previously, we substitute all occurrences of any target word (denoted as multiple-swap) and found that frequent words in general have smaller sensitivity rankings, i.e., higher sensitivity scores. The sensitivity ranking and word frequency are correlated with a Pearson coefficient of -0.62 for tf-idf as shown in \autoref{fig:ts} and Appendix \ref{apd:second}. The Pearson coefficient is -0.60 for BioClinicalBERT, as shown in \autoref{fig:bs} and Appendix \ref{apd:second}. Thus, the metric is biased if we compare sensitivities of tokens with diverging frequencies.

To tackle this problem, we substitute the \textit{first} occurrence of the target word (called ``one-swap" for brevity) rather than multiple-swap. Figure \ref{fig:ts} and \ref{fig:bs} shows that with this method, the sensitivity ranking are distributed more evenly across word frequency. Nevertheless, it should be acknowledged that under rare circumstances, swapping the first occurrence could introduce a small bias when its particular context lend higher importance to this occurrence than others.

To get some insights about the correlation between word frequency and sensitivity, we consider a toy example of a linear classifier using the TD-IDF embedding. Suppose we have an input 
$$x=(``\text{hi}",``\text{there}")$$

and a perturbed input with respect to the token of interest $u$=``there":
$$g(x)=(``\text{hi}",``\text{hi}")$$
The tf-idf+linear classifier is defined as
$$f_{\text{tf-idf+linear}}(x) = \text{softmax}(W\phi(x)).$$
Here $\phi$ is the tf-idf embedding function with two vocab: "hi" and "there":
$$\phi(x)=
\begin{bmatrix}
\text{tf}(\text{``hi"},x)\log(\frac{|X|}{\text{df}(\text{``hi",x})}) \\
\text{tf}(\text{``there"},x)\log(\frac{|X|}{\text{df}(\text{``there",x})})
\end{bmatrix},
$$

where term frequency tf($w$,$x$) is the number of $w$ in a note $x$ and document frequency df is the number of notes $x\in X$ that contain token $w$.

The key insight is that a higher term frequency of the token of interest tf($u,x$) would lead to a larger change in $L_1$ norm of the perturbed tf-idf embedding. For example, both ``hi" and ``there" appears once in $x$, so
$$\phi(x)=
\begin{bmatrix}
\log(\frac{|X|}{\text{df}(\text{``hi",x})}) \\
\log(\frac{|X|}{\text{df}(\text{``there",x})})
\end{bmatrix}.
$$

After perturbation, ``there" completely disappear while there are 2 ``hi"s, so
$$\phi(g(x))=
\begin{bmatrix}
2\log(\frac{|X|}{\text{df}(\text{``hi",x})}) \\
0
\end{bmatrix}.
$$

The difference in tf-idf embedding is
$$
\|\phi(x)-\phi(g(x))\|_1 
$$
$$=\log(\frac{|X|}{\text{df}(\text{``hi",x})}) + \log(\frac{|X|}{\text{df}(\text{``there",x})})$$

More generally, given a word of interest $u$ with term frequency $n$, the difference in tf-idf embeddings scales with $n$:
\begin{align*}
&\|\phi(x)-\phi(g(x))\|_1 \\ 
&= n\log(\frac{|X|^2}{\text{df}(\text{``hi",x})\text{df}(\text{``there",x})}) \\
&\implies\|\phi(x)-\phi(g(x))\|_1\propto n
\end{align*}


In our toy example, since $f$ is linear, we know the difference in predicted probability would be larger if the perturbed word of interest has a larger term frequency $n$:
\begin{equation}\label{eq:tf_prob}
    |f(x)-f(g(x))|\propto \|\phi(x)-\phi(g(x))\|_1 \propto n
\end{equation}

On average, a token with a higher word frequency (meaning that it appears more often in the dataset) have a higher term frequency in each note. By \autoref{eq:tf_prob}, such high-frequency token has a higher sensitivity score after perturbation.

\subsection{Perturbation Filter}
\label{appen:C2}
Each filter in uniform perturbations replaces the token of interest $u$ with another token uniformly sampled from the vocabulary of BioClinical Bert.
\begin{equation*}
    U = \{h:h(u)=w',w'\in W_{\text{uniform}}\}
\end{equation*}

Each filter in 1-gram perturbations replaces the token of interest $u$ with one of the five most frequent tokens from the subset $X$ that contains $u$.
\begin{equation*}
     G = \{h:h(u)=w',w'\in W_{\text{1-gram}}\}
\end{equation*}

Each filter in context perturbations replaces the token of interest $u$ with one of the five most likely predictions according to non-finetuned BioClinicalBERT's MLM probability. We specifically used the \textit{non-finetuned} BioClinicalBERT to avoid using the same model for both prediction and assessment. 
\begin{equation*}
    O= \{h:h(u)=w',w'\in W_{\text{context}}\}
\end{equation*}

\subsection{Spearman Rank Correlation}

\label{appen:C3}
\begin{equation}\label{my_eighth_eqn}
 r_s = 1-6 \cdot \frac{\sum D^2}{n(n^2-1)}
\end{equation}

We use the Spearman rank correlation \citep{spearman_proof_1904} to quantify the divergence between the model ranking and the manual ranking. In \autoref{my_eighth_eqn}, $D$ is the difference between ranks, and $n$ is the number of pairs of data.
\section{Tables}\label{apd:second}
The sensitivity ranking and word frequency are negatively correlated for both the tf-idf model and BioClinicalBERT when we swap multiple occurrences. 

Despite similar AUROC scores, BioClinicalBERT is more sensitive to disease names compared to general words, while tf-idf+xgb displays a more irregular distribution of sensitivity.  

\autoref{tab:table3} and \autoref{tab:table4} display the Sensitivity Score of 10 words with different frequencies with respect to the language model and the tfidf+xgboost model using multiple swaps of occurrences. The one-swap counterparts are shown in \autoref{table:table5} and  \autoref{table:table6}. To assess the models' alignment with professional opinions, \autoref{tab:49_words_1} lists the token sets and words' relative significance ranking.

\clearpage

\begin{table}[h]
\small
  \begin{tabular}{llll}
  \toprule
  \bfseries word & \bfseries test\_fre &\bfseries $\overline{{d_{f}}}(g_{u,H})(X)$ & \bfseries rank\\
  \midrule
  cancer & 1912 & 0.033884  & 1 \\
  mg & 58910 & 0.027513 & 2 \\
  colon & 395 & 0.020076 & 3 \\
  expired & 1234 & 0.018921 & 4 \\
  deceased & 399 & 0.017083 & 5 \\
 heparin & 1898 & 0.014974 & 6\\
  died & 1871 & 0.012532  & 7 \\
 father & 1890 & 0.007779 & 8 \\
  mother& 1897 & 0.006713 & 9 \\
  mouthwash & 78 & 0.005541 & 10\\
  regimen & 395 & 0.004930 & 11\\
  congenital & 78 & 0.002269 & 12\\
  thinner & 78 &  0.002439 & 13 \\
  \bottomrule
  \end{tabular}
  \caption{The language model's Sensitivity Score of 10 words within different frequency range (multiple-swap)}
  \label{tab:table3}
\end{table}

\begin{table}[h]
\small
  {\begin{tabular}{llll}
  \toprule
  \bfseries word & \bfseries test\_fre &\bfseries $\overline{{d_{f}}}(g_{u,H})(X)$ & \bfseries rank\\
  \midrule

  expired & 1234 & 0.019003 & 1 \\
  mg & 58910 & 0.007416 & 2 \\
   heparin & 1898 & 0.000583 & 3\\
  mouthwash & 78 & 0.005541 & 4\\
 deceased & 399 & 0.017083 & 5 \\
  died & 1871 & 0.012532  & 6 \\
   cancer & 1912 & 0.033884  & 7 \\
  regimen & 395 & 0.004930 & 8\\
  colon & 395 & 0.020076 & 9 \\
  mother& 1897 & 0.006713 & 10 \\
 father & 1890 & 0.007779 & 11 \\
  thinner & 78 &  0.002439 & 12 \\
  congenital & 78 & 0.002269 & 13\\
  \bottomrule
  
  \end{tabular}}
\caption{The tf-idf + xgboost model's Sensitivity Score of 10 words within different frequency range (multiple-swap)}
\label{tab:table4}
\end{table}

\begin{table}[h]
\small  
  {\begin{tabular}{llll}
  \toprule
  \bfseries word & \bfseries test\_fre &\bfseries $\overline{{d_{f}}}(g_{u,H})(X)$ & \bfseries rank\\
  \midrule
  hypoglycemia & 168 &  0.025779 &1\\
  fall & 789 & 0.021095 & 2\\
  ulcer & 358 & 0.011134 & 3\\
  prematurity & 164 & 0.010929 & 4\\
  arthritis & 158 & 0.009826 & 5\\
  father & 1890 & 0.007510  & 6\\
  mother & 1897 & 0.006404  & 7\\
  patient & 7900 & 0.005129  & 8\\
  blood & 6565 & 0.004980 & 9\\
  labor &16 & 0.003916 & 10\\
  vaccination & 78 & 0.001465 &11 \\
  \bottomrule
  
  \end{tabular}}
  \caption{BioClinicalBERT's Sensitivity Score of a list of ailments and words for comparison (one-swap)}
  \label{table:table5}
\end{table}

\begin{table}[h]  
\small
  {\begin{tabular}{llll}
  \toprule
  \bfseries word & \bfseries test\_fre &\bfseries $\overline{{d_{f}}}(g_{u,H})(X)$ & \bfseries rank\\
  \midrule
  fall & 789 &  0.003331 & 1\\
  blood & 6565 & 0.000481 & 2\\
  hypoglycemia & 168 & 0.000426 &3\\
  patient & 7900 & 0.000342 & 4\\
  ulcer & 358 &  0.000319 & 5\\
  prematurity & 164 &  0.000249 & 6\\
  arthritis & 158 & 0.000248 & 7\\
  mother & 1897 & 0.000248 & 8\\
  father & 1890 & 0.000239 & 9\\
  vaccination & 78 & 0.000024 &10 \\
  labor &16 & 0.000000 & 11\\
  \bottomrule
  
  \end{tabular}}
  \caption{tf-idf+xgb's Sensitivity Score of a list of ailments and words for comparison (one-swap)}
  \label{table:table6}
\end{table}
\clearpage
\begin{table}[h]
\small
  \begin{tabular}{llll}
  \toprule
  \bfseries word & \bfseries manual &\bfseries language & \bfseries  tf-idf\\
  \midrule
    chemotherapy & 1.0 & 15 & 25 \\
    hypoglycemia & 2.0 & 3 & 12 \\
    tumor & 3.0 & 19 & 40 \\
    overdose & 3.0 & 1 & 2 \\
    dementia & 3.0 & 2 & 16 \\
    anticoagulation & 6.0 & 36 & 36 \\
    delirium & 6.0 & 26 & 17 \\
    debridement & 6.0 & 7 & 10 \\
    arrhythmia & 6.0 & 5 & 20 \\
    pancreatic & 6.0 & 4 & 35 \\
    amputation & 6.0 & 10 & 23 \\
    fall & 6.0 & 6 & 3 \\
    cardiovascular & 13.0 & 17 & 45 \\
    neurosurgery & 13.0 & 31 & 33 \\
    diabetes & 13.0 & 39 & 24 \\
    ablation & 16.0 & 11 & 15 \\
    expired & 16.0 & 21 & 1 \\ 
    dehydrated & 16.0 & 25 & 41 \\
    palpitations & 16.0 & 8 & 28 \\
    obesity & 16.0 & 41 & 34 \\
    wheeze & 21.0 & 14 & 27 \\
    vaccination & 21.0 & 47 & 48 \\
    arthritis & 21.0 & 22 & 30 \\
    pain & 21.0 & 27 & 32 \\
    dysfunction & 21.0 & 23 & 8 \\
  \bottomrule
  \end{tabular}
  \caption{Sensitivity Score rankings of 49 hand-chosen words for model comparison (one-swap). The language model's sensitivity ranking aligns better with the clinicians' manual rankings. (first half)}
  \label{tab:49_words_1}
\end{table}

\begin{table}[h]
\small
  \begin{tabular}{llll}
  \toprule
  \bfseries word & \bfseries manual &\bfseries language & \bfseries  tf-idf\\
  \midrule
    urinary & 26.0 & 24 & 5 \\
    faint & 26.0 & 46 & 42 \\
    refills & 26.0 & 9 & 19 \\
    immunizations & 26.0 & 38 & 39 \\
    blood & 26.0 & 34 & 9 \\
    family & 26.0 & 35 & 37 \\
    diarrhea & 32.0 & 16 & 22 \\
    female & 32.0 & 18 & 44 \\
    prescribed & 32.0 & 30 & 7 \\
    medication & 32.0 & 45 & 29 \\
    electrolytes & 32.0 & 40 & 43 \\
    allergies & 32.0 & 12 & 46 \\
    aspirin & 32.0 & 13 & 6 \\
    increase & 32.0 & 48 & 4 \\
    tylenol & 40.0 & 20 & 13 \\
    care & 40.0 & 37 & 26 \\
    benign & 40.0 & 29 & 38 \\
    mother & 40.0 & 28 & 31 \\
    cartridge & 40.0 & 44 & 14 \\
    labor & 40.0 & 43 & 49 \\
    moderate & 40.0 & 49 & 47 \\
    tablet & 40.0 & 33 & 21 \\
    mg & 40.0 & 42 & 11 \\
    patient & 40.0 & 32 & 18 \\
  \bottomrule
  \end{tabular}
  \caption{Sensitivity Score rankings of 49 hand-chosen words for model comparison (one-swap). The language model's sensitivity ranking aligns better with the clinicians' manual rankings. (second half)}
  \label{tab:49_words_2}
\end{table}
\end{document}